\documentclass[sigconf,nonacm]{aamas}

\usepackage{amsmath}
\usepackage{algorithm}
\usepackage{algpseudocode}
\usepackage{colortbl}
\usepackage{tabularx}
\definecolor{TableGroup}{HTML}{EBEDF0}
\definecolor{TableStripe}{HTML}{F7F7F8}
\definecolor{TableHighlight}{HTML}{EEE9F7}

\graphicspath{{figures/}}

\setcopyright{none}
\acmConference{}{}{}{}
\acmDOI{}
\acmISBN{}
\submissionType{}

\title[What Should an Agent Forget?]{\texorpdfstring{What Should an Agent Forget?\\Separating What Is Stored from What Is Used}{What Should an Agent Forget? Separating What Is Stored from What Is Used}}
\author{Yuhang Li}
\authornote{Both authors contributed equally to this work.}
\affiliation{\institution{Beihang University}\city{Hangzhou}\country{China}}
\author{Yuchen Li}
\authornotemark[1]
\affiliation{\institution{East China Normal University}\city{Shanghai}\country{China}}
\renewcommand{\shortauthors}{Yuhang Li and Yuchen Li}
\hypersetup{keeppdfinfo}

\begin{abstract}
Persistent language agents need stored experience to remain available across time, while each answer requires evidence suited to a particular question. A superseded fact can mislead a current-state answer and still be essential for a historical query. We present RD-Forget, a training-free framework that separates what an agent stores from what it uses. A retained source archive preserves observations, and a query-conditioned memory view controls their influence on the current answer. A frozen language-model curator extracts relevant evidence, groups facts into semantic slots, and preserves the relations needed for multi-hop reasoning. Same-slot replacement links suppress superseded values in current-state contexts, while intent-aware retrieval makes earlier evidence eligible again. A rate-distortion formulation guides construction of the answer-time view within a memory budget. Experiments span conversational memory, knowledge updating, fact consolidation, long-context reasoning, and personalization under a shared answering pipeline. The results associate accurate answers with both query-relevant evidence construction and control over obsolete alternatives. Configurations without forgetting or query conditioning have the largest score deficits, while slot grouping, historical access, and relation preservation contribute complementary functions. Retaining history while selectively controlling its use offers a practical way to accommodate changing facts and future questions.
\end{abstract}

\keywords{Language agents, persistent memory, selective forgetting, query-conditioned retrieval}

\newcommand{\RD}{RD-Forget}
\newcommand{\tok}{\operatorname{Tok}}
\newcommand{\slot}{\operatorname{slot}}
\newcommand{\status}{\operatorname{status}}
\newcommand{\rdActive}{\mathsf{ACTIVE}}
\newcommand{\rdSuperseded}{\mathsf{SUPERSEDED}}

\begin{document}
\maketitle
\pagestyle{plain}\thispagestyle{plain}

\begin{figure*}[t]
  \centering
  \includegraphics[width=\textwidth]{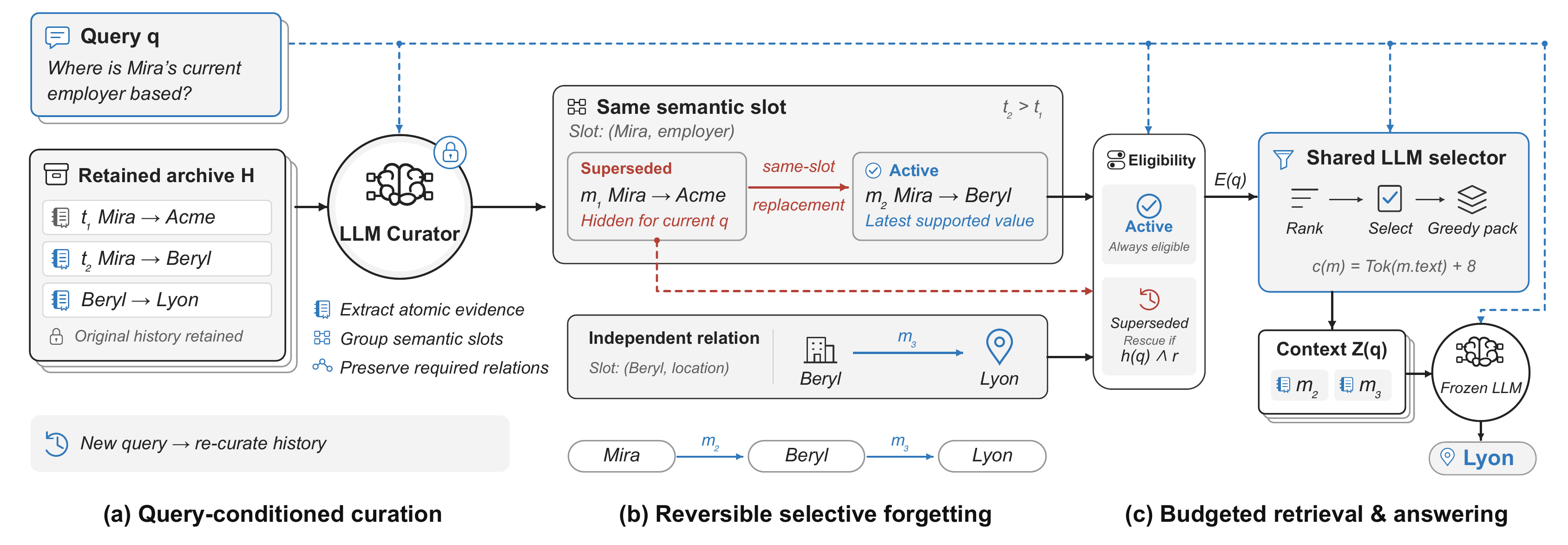}
  \caption{Separating storage and use in \RD. The archive retains source observations; the query-conditioned view determines the evidence used for an answer. Same-slot replacement suppresses the earlier employer while preserving the independent location relation. Historical intent and the rescue switch jointly make superseded entries eligible again. The shared selector and packer construct the answer context, illustrated by the Mira--Beryl--Lyon chain.}
  \label{fig:method}
  \Description{Three stages of RD-Forget: query-conditioned curation, selective forgetting, and budgeted retrieval and answering. Mira's employer changes from Acme to Beryl, while Beryl's location remains Lyon. The old employer is superseded, and the current employer and its location stay active. Superseded evidence is eligible only when both historical intent and rescue are enabled. A shared selector and greedy packer produce the context used by a frozen language model to answer Lyon. The archive remains available for curation when a new query arrives.}
\end{figure*}

\section{Introduction}
Persistent language agents face two decisions about memory: what experience should remain stored, and what evidence should influence the next answer. These decisions diverge when the world changes. An earlier employment record can mislead a question about a user's current employer while correctly answering a question about a previous job. The observation remains useful in storage even when its use must change. Separating retention from answer-time access allows an agent to preserve history while excluding obsolete alternatives from a current-state response.

External memory gives language agents access to information beyond a single interaction. Generative Agents accumulate observations and reflections~\citep{park2023generative}; MemGPT manages information across memory tiers~\citep{packer2023memgpt}; A-MEM builds linked notes that evolve with new observations~\citep{xu2025amem}. ACE and ReasoningBank further develop the accumulation and reuse of task experience~\citep{zhang2025ace,ouyang2025reasoningbank}. As experience accumulates, memory management must connect what remains available in storage to what is appropriate for the current answer.

The use of a fact also depends on its relation to other evidence. A new employer supersedes an old employer within one relation, while an older fact about the new employer's location may supply a necessary reasoning step. Recency alone does not distinguish a competing value from a complementary relation. A compact general-purpose memory can omit a relation whose value becomes clear only when the question arrives. Useful forgetting must therefore respect both the scope of an update and the evidence chain required by the query.

We make the separation between stored history and used evidence explicit in \RD. A source archive retains observations, and a query-dependent memory view controls their influence on the next answer. A frozen language-model curator identifies relevant facts and their semantic slots. Same-slot replacement links suppress superseded values for current-state questions, while historical intent allows earlier evidence to become eligible again. The curator preserves the relations needed to derive an answer, including complementary facts that remain useful after an update. Figure~\ref{fig:method} shows how the retained archive supports these changing views.

We formulate construction of the answer-time view through a rate--distortion trade-off: selected evidence should preserve answer utility within a representation budget. The archive supplies observations that can be reconsidered when the question changes. Curation determines relevance, supersession determines revision eligibility, and budgeted selection forms the final context. Forgetting thus has a query-local scope: an observation can remain stored while its influence on a particular answer is suppressed.

We evaluate five task suites with four language models. The tasks cover conversational memory, knowledge updates, fact consolidation, long-context reasoning, and personalization. On fact consolidation, gains over the better-performing baseline range from 11.00 to 26.00 percentage points across the four backbones. In the matched Luna ablations, the configuration without forgetting scores lowest, followed by the query-unconditioned configuration. These results connect the distinction between storage and use to which facts enter the view and which revisions remain eligible for answering.

\section{Related Work}
\subsection{Selective forgetting and consolidation}
Forgetting can act on the information retained in working context or on the organization of persistent memory. MemAct learns context insertion and deletion actions through reinforcement learning~\citep{zhang2025memact}. FadeMem combines adaptive decay, relevance and access signals, conflict resolution, and memory fusion~\citep{wei2026fademem}. \RD\ conditions forgetting on the current question and the semantic scope of a replacement. An old fact can leave the current answer context while remaining accessible for historical use. This connects forgetting to changing evidence requirements across queries.

\subsection{External memory and context organization}
Generative Agents store observations, synthesize reflections, and retrieve memories to guide behavior~\citep{park2023generative}. MemGPT uses virtual context management to move information between memory tiers~\citep{packer2023memgpt}. A-MEM constructs notes with contextual attributes and links, allowing later observations to update earlier memory organization~\citep{xu2025amem}. These systems provide structures for retaining and accessing experience. \RD\ focuses on constructing the used evidence from retained history, using query intent and semantic slots to distinguish replacement from complementary information.

\subsection{Experience-based adaptation}
ACE maintains evolving playbooks through generation, reflection, and incremental curation, with structured delta updates and a grow-and-refine process~\citep{zhang2025ace}. ReasoningBank extracts reusable strategies from successful and failed experiences and combines memory with test-time computation through MaTTS~\citep{ouyang2025reasoningbank}. Their update policies motivate the comparison between editing existing memory and accumulating additional entries.

\subsection{Evaluating evolving memories}
LoCoMo evaluates long-term conversational memory~\citep{maharana2024locomo}. LongMemEval distinguishes extraction, multi-session reasoning, temporal reasoning, knowledge updates, and abstention~\citep{wu2024longmemeval}. MemoryAgentBench includes conflict resolution through fact consolidation~\citep{hu2025memoryagentbench}. BEAM adds long conversations with knowledge-update, contradiction, temporal, and preference probes~\citep{tavakoli2025beam}; PersonaMem evaluates responses aligned with evolving user profiles~\citep{jiang2025personamem}. Together, these tasks expose different consequences of forgetting: obsolete values can corrupt current answers, while discarded history or relations can make later questions unanswerable.

\section{Separating Storage from Use}
\subsection{Stored history and used evidence}
Let $H=(e_1,\ldots,e_T)$ be the stored source history available to query $q$. Observations can carry session identifiers, timestamps, or sequence indices. Evidence used for answering forms a memory view $Z(q)$ derived from $H$ and passed to a frozen answer model $g_\theta$. \RD\ constructs this view through evidence extraction, replacement relationships, and eligibility decisions. These operations change the materialized entries available for an answer; their source observations remain in $H$. A later query can therefore produce a different view from the retained history.

A rate--distortion formulation applies the representation cost to the answer-time view and measures distortion through loss of answer utility~\citep{cover2005elements}:
\begin{align}
 Z^*(q) &\in \arg\min_{Z\subseteq C(q)} D_q(Z), \label{eq:objective}\\
 \text{subject to}\qquad &\sum_{m\in Z}c(m)\leq B, \nonumber
\end{align}
where $C(q)$ is the query-dependent candidate pool and
\begin{equation}
 D_q(Z)=\mathbb{E}\big[d(Y,g_\theta(q,Z))\mid q,H\big]
\end{equation}
measures task loss $d$ against the correct answer $Y$. The query determines which omissions are costly: forgetting a replaced value may help a current-state answer, while forgetting the same observation can prevent a historical answer. \RD\ implements the selection through language-model curation, status-based eligibility, and greedy packing.

The experiments use the memory accounting rule
\begin{equation}
 c(m)=\tok(m.\mathrm{text})+8,\qquad B=2{,}048,
 \label{eq:cost}
\end{equation}
where $\tok$ uses the \texttt{o200k\_base} tokenizer and the constant covers per-entry overhead. The budget governs the selected view $Z(q)$. Reading stored history, query text, and other prompt instructions contributes separately to model input.

\subsection{Building the query-dependent view}
\label{sec:curation}
The curator extracts active evidence and facts identified as replaced from $H$ for $q$. Each item contains an atomic statement, a proposed fact slot, a source-evidence string, and an optional validity time. The slot has the form \texttt{subject|relation|scope}. It makes replacement specific to a relation: changing a person's employer updates their employment slot, while a fact about the employer's location belongs to another slot.

The curator is prompted to extract the smallest complete evidence set needed by the question. For a multi-hop query, it retains each relation needed to reach the answer and resolves updates within each hop. In Figure~\ref{fig:method}, the current employment relation and the employer-location relation jointly support the answer. Suppressing the previous employer removes a competing value while preserving that chain.

Temporal interpretation follows the history format. In MAB-FC, larger sequence numbers identify later updates within a relation slot. AMB-Text exposes the conversation through the query's session boundary. LME-KU supplies oracle sessions and a query date; current-state questions use the latest explicit update in those sessions. Within a slot, the curator distinguishes a replacement from a related observation that can coexist with the current value.

\begin{figure*}[t]
  \centering
  \includegraphics[width=\textwidth]{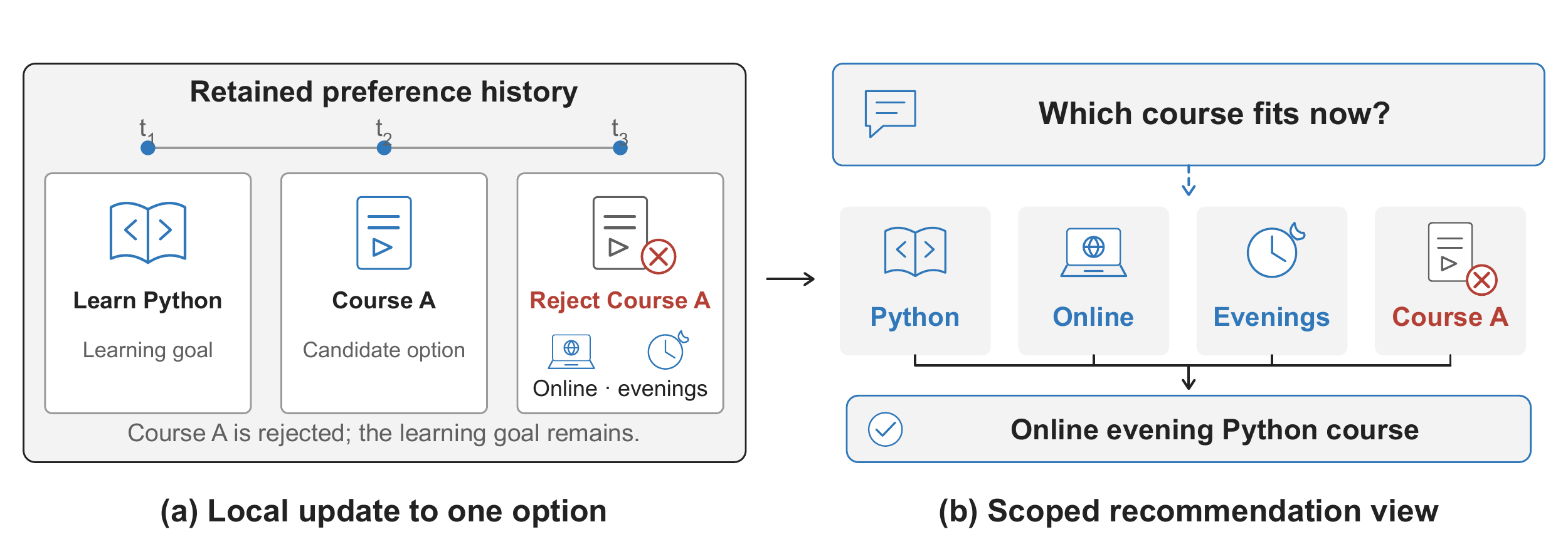}
  \caption{Scope-aware preference curation, shown through a constructed example. (a) A user wants to learn Python, considers Course A, then rejects that option and requests online evening study. (b) The recommendation view keeps the learning goal, applies the new format and schedule constraints, and excludes Course A. Rejecting an option does not imply rejecting the broader interest. Icons represent evidence and constraints, not measured execution steps.}
  \label{fig:intent-views}
  \Description{A timeline uses book and course icons to show learning Python, considering Course A, and rejecting Course A while requesting online evening study. The recommendation view combines a book, laptop, evening clock, and crossed-out Course A card into an online evening Python-course recommendation. The rejection is local to Course A.}
\end{figure*}

The intent-aware extension also handles questions that need several states at once. Contradiction-resolution questions retain the relevant claims, evolution questions preserve turning points, and recommendation questions combine useful preferences with later constraints. Scope remains local to the proposition being updated. Rejecting a particular course, for example, need not replace a broader preference for learning that skill. Figure~\ref{fig:intent-views} illustrates how a recommendation view retains the learning goal while incorporating an option-specific rejection and updated study constraints.

The curator has access to the question and source evidence; gold answers and judge decisions are used during evaluation. If curation produces no usable entries, a lexical fallback returns up to 20 nonempty source lines ranked by overlap with the question.

\subsection{Controlling use through supersession}
\label{sec:materialization}
The materializer normalizes slot strings by stripping whitespace and lowercasing them, then creates active entries before processing proposed replacements. The last active entry in each slot supplies the replacement target. Let $m_o$ be an older fact and $m_n$ an active entry. A superseded record is created when
\begin{equation}
 \slot(m_o)=\slot(m_n),\quad
 \status(m_n)=\rdActive,
 \label{eq:replacement}
\end{equation}
and both entries have content. The older entry's \texttt{superseded\_by} field points to $m_n$. Rows without an active same-slot replacement are omitted from the materialized memory, while their source observations remain in $H$.

This ordering ties each supersession decision to a concrete replacement. It separates two reasons for excluding a fact: the fact can be irrelevant to the present query, or it can have been explicitly replaced within its slot. The latter receives a stored relationship that subsequent retrieval can use. With forgetting disabled, the older fact stays active and its supersession link is cleared. With slot grouping disabled, older conflicts remain separate active entries.

\subsection{Reusing stored history}
\label{sec:rescue}
Let $A$, $D$, and $S$ denote materialized entries with active, deprecated, and superseded status, and let $r$ indicate whether archive rescue is enabled. Retrieval forms the eligible pool
\begin{align}
 E(q) = A
 &\;\cup\; \begin{cases}D,&r=1,\\\varnothing,&r=0,\end{cases} \nonumber\\
 &\;\cup\; \begin{cases}S,&r=1\text{ and }h(q)=1,\\\varnothing,&\text{otherwise},\end{cases}
 \label{eq:eligibility}
\end{align}
where $h(q)$ identifies a historical or revision-chain request. Its input convention is specified in Section~\ref{sec:execution}. The selector is instructed to use deprecated evidence when it is relevant and lacks an active substitute. The query-conditioned curator produces active and superseded entries; the incremental memory interface also supports deprecated entries.

Older evidence can return through two paths. A new query can cause the curator to extract it from $H$ into a new memory view. Within an existing materialized view, rescue makes a non-active entry eligible for selection. The \texttt{archive\_rescues} counter measures selected entries on this second path. A historical fact extracted directly into active memory uses the first path.

These paths preserve access to stored history while changing what is used. In Figure~\ref{fig:method}, a current employer query uses Beryl, a historical query can recover Acme, and an evolution query can retain both states. The query changes the evidence view drawn from the archive.

\subsection{Packing the answer-time view}
Eligible entries are ranked by lexical overlap with $q$, with an additional score of 0.4 for active entries. The candidate collector admits the first ranked entry and adds further entries within a 10,000-token serialization target. A language-model selector, shared across the compared policies, returns an ordered list of entry identifiers. Candidate order serves as the fallback when no usable identifiers are returned.

The packer traverses this ordering and includes an entry when its cost fits the remaining memory budget. Oversized entries are skipped. This gives $\sum_{m\in Z}c(m)\leq B$ for the selected memory. Packing operates on individual entries, while the curator's chain instruction supplies the complementary relations available for selection. The answer model receives $q$ and the selected entries with their identifiers and statuses.

Algorithm~\ref{alg:rd} summarizes the procedure. With $N$ materialized entries, deterministic ranking takes $O(N\log N)$ comparisons and packing is linear in the returned ordering.

\begin{algorithm}[t]
\caption{From stored history to an answer in RD-Forget}
\label{alg:rd}
\small
\begin{algorithmic}[1]
\Require Available history $H$, query $q$, budget $B$, rescue switch $r$
\State $(A_0,F_0)\gets\textsc{Curate}_{\theta}(H,q)$
\State $M\gets\textsc{MaterializeActive}(A_0)$
\For{older item $f\in F_0$}
  \If{an active same-slot replacement $m$ exists}
    \State append $f$ with status $\rdSuperseded$ to $M$
    \State link $f.\mathrm{superseded\_by}\gets m.\mathrm{id}$
  \EndIf
\EndFor
\If{$M$ is empty}
  \State $M\gets\textsc{LexicalSourceFallback}(H,q)$
\EndIf
\State $E\gets\textsc{Eligible}(M,q,r)$ using Eq.~\eqref{eq:eligibility}
\State $C\gets\textsc{RankAndCollect}(E,q)$
\State $\pi\gets\textsc{Select}_{\theta}(C,q)$
\State use candidate order if $\pi$ has no usable identifiers
\State $Z\gets[\ ];\quad b\gets0$
\For{entry $m$ in $\pi$}
  \If{$b+c(m)\leq B$}
    \State append $m$ to $Z$; $b\gets b+c(m)$
  \EndIf
\EndFor
\State \Return $g_{\theta}(q,Z)$
\end{algorithmic}
\end{algorithm}

\section{Experimental Setup}
\label{sec:protocol}
\subsection{Task suites}
The main evaluation covers three question-answering suites. \textbf{AMB-Text} contains 86 text questions from the first LoCoMo conversation~\citep{maharana2024locomo} as distributed with AgentMemoryBench~\citep{hu2025memoryagentbench}. We exclude category 5, multimodal questions, and questions without a session boundary. Each query receives the conversation through its boundary. \textbf{LME-KU} contains the 78 knowledge-update questions in the LongMemEval oracle data~\citep{wu2024longmemeval}. \textbf{MAB-FC} contains 100 question--answer pairs from the \texttt{factconsolidation\_mh\_6k} source in MemoryAgentBench's Conflict\_Resolution split~\citep{hu2025memoryagentbench}. These questions share one fact history.

The three suites comprise 264 tasks per method and 792 method--task combinations per model. Selection follows source order and fixed index ranges. The query-intent evaluation adds 150 BEAM questions and 579 PersonaMem questions per method, excluding the first ten development questions in each benchmark. Section~\ref{sec:extended} describes these two suites and their scoring rules.

\subsection{Models and execution}
\label{sec:execution}
We evaluate Qwen3.5-flash, GPT-5.6-Luna, MiniMax-M2.5, and Kimi-K2.5. Within each run, curation, selection, answering, and model-based evaluation use the same model family and client configuration. We therefore treat model-judged scores as within-protocol evaluation rather than as fully independent external judgments. 
Available query dates and task metadata accompany curation and selection; $h(q)$ combines lexical cues with task-type labels for contradiction resolution, temporal reasoning, and preference history. These auxiliary inputs are implicit in Algorithm~\ref{alg:rd}.

The main-suite and matched-Full protocol allows up to 4,096 output tokens for curation and 256 for each selection, answer, and judge call. The query-intent protocol uses caps of 8,192 and 2,048, respectively, and an 8,000-token source chunk target. 

\subsection{Scoring and aggregation}
AMB-Text and LME-KU report mean binary semantic judgments against the gold answer. MAB-FC reports rule accuracy from normalized substring matching. The rule lowercases and tokenizes text, converts number words from zero through twelve to digits, and accepts a nonempty gold substring in the answer or an answer of at least three characters within a gold string. Accuracy is computed over completed evaluations.

\begin{table*}[t]
  \centering
  \caption{Main comparison (\%). Mean equally weights the three displayed benchmark scores. Bold marks column bests within each model; purple identifies \RD.}
  \label{tab:main}
  \small
  \setlength{\tabcolsep}{4pt}
  % Generated from analysis/reported_results.json; user summary takes precedence.
\begin{tabularx}{\textwidth}{@{}l*{4}{>{\raggedleft\arraybackslash}X}@{}}
\toprule
\textbf{Memory policy} & \multicolumn{3}{c}{\textbf{Answer accuracy (\%)} $\uparrow$} & \textbf{Mean} $\uparrow$ \\
\cmidrule(lr){2-4}
 & AMB-Text & LME-KU & MAB-FC &  \\
\midrule
\rowcolor{TableGroup}\multicolumn{5}{@{}l}{\textit{Qwen3.5-flash}} \\
ACE & 52.33 & 60.26 & 34.00 & 48.86 \\
\rowcolor{TableStripe}ReasoningBank & 54.65 & 60.26 & 23.00 & 45.97 \\
\rowcolor{TableHighlight}\textbf{RD-Forget (ours)} & \textbf{74.42} & \textbf{74.36} & \textbf{51.00} & \textbf{66.59} \\
\midrule
\rowcolor{TableGroup}\multicolumn{5}{@{}l}{\textit{GPT-5.6-Luna}} \\
ACE & 67.44 & 83.33 & 51.00 & 67.26 \\
\rowcolor{TableStripe}ReasoningBank & 75.58 & 71.79 & 46.00 & 64.46 \\
\rowcolor{TableHighlight}\textbf{RD-Forget (ours)} & \textbf{89.53} & \textbf{93.59} & \textbf{77.00} & \textbf{86.71} \\
\midrule
\rowcolor{TableGroup}\multicolumn{5}{@{}l}{\textit{MiniMax-M2.5}} \\
ACE & 69.77 & 61.54 & 42.00 & 57.77 \\
\rowcolor{TableStripe}ReasoningBank & 62.79 & 51.28 & 46.00 & 53.36 \\
\rowcolor{TableHighlight}\textbf{RD-Forget (ours)} & \textbf{87.21} & \textbf{87.18} & \textbf{57.00} & \textbf{77.13} \\
\midrule
\rowcolor{TableGroup}\multicolumn{5}{@{}l}{\textit{Kimi-K2.5}} \\
ACE & 81.40 & 70.51 & 41.00 & 64.30 \\
\rowcolor{TableStripe}ReasoningBank & 80.23 & 62.82 & 45.00 & 62.68 \\
\rowcolor{TableHighlight}\textbf{RD-Forget (ours)} & \textbf{82.56} & \textbf{76.92} & \textbf{59.00} & \textbf{72.83} \\
\bottomrule
\end{tabularx}

\end{table*}

\subsection{Ablations}
\label{sec:ablations}
The matched Full batch and five variants use the same 264 tasks. $-$FORGET keeps identified older revisions active. $-$RESCUE excludes non-active entries from retrieval. $-$SLOT removes shared slot grouping and retains older conflicts as separate active facts. $-$CLOSURE instructs the curator to keep directly matched evidence without expanding the complete multi-hop chain. The selector retains its common instruction to preserve necessary links.

$-$QC hides the question from the curator and builds a general-purpose memory per source. The downstream selector and the empty-curation fallback still receive the question. These variants compare complete curation configurations: switches can change the upstream extracted evidence as well as entry eligibility. 

\begin{table*}[t]
  \begin{minipage}[t]{0.48\textwidth}
    \centering
    \caption{Luna ablations (\%). The shaded matched Full is a separate batch from Table~\ref{tab:main}. Bold marks column bests, including ties.}
    \label{tab:ablations}
    \small
    \setlength{\tabcolsep}{3pt}
    % Generated from analysis/reported_results.json; user summary takes precedence.
\begin{tabularx}{\linewidth}{@{}l*{3}{>{\raggedleft\arraybackslash}X}@{}}
\toprule
Variant & AMB-Text & LME-KU & MAB-FC \\
\midrule
\rowcolor{TableHighlight}\textbf{matched Full} & \textbf{91.86} & \textbf{93.59} & \textbf{74.00} \\
$-$FORGET & 68.60 & 60.26 & 51.00 \\
$-$RESCUE & 88.37 & 91.03 & 65.00 \\
$-$SLOT & 86.05 & 89.74 & 64.00 \\
$-$CLOSURE & 87.21 & 89.74 & 71.00 \\
$-$QC & 77.91 & 82.05 & 56.00 \\
\bottomrule
\end{tabularx}

  \end{minipage}\hfill
  \begin{minipage}[t]{0.48\textwidth}
    \centering
    \caption{Query-intent evaluation (\%). BEAM averages rubric scores; PersonaMem uses single-option accuracy. Bold marks row bests.}
    \label{tab:extended}
    \small
    \setlength{\tabcolsep}{3pt}
    % Generated from analysis/reported_results.json; user summary takes precedence.
\begin{tabularx}{\linewidth}{@{}l*{3}{>{\raggedleft\arraybackslash}X}@{}}
\toprule
Model & RD & ACE & RB \\
\midrule
\rowcolor{TableGroup}\multicolumn{4}{@{}l}{\textit{BEAM}} \\
Qwen3.5-flash & \textbf{50.21} & 32.84 & 34.75 \\
\rowcolor{TableStripe}GPT-5.6-Luna & \textbf{70.58} & 56.97 & 53.47 \\
MiniMax-M2.5 & \textbf{42.39} & 31.85 & 33.92 \\
\rowcolor{TableStripe}Kimi-K2.5 & \textbf{57.83} & 42.16 & 41.83 \\
\midrule
\rowcolor{TableGroup}\multicolumn{4}{@{}l}{\textit{PersonaMem}} \\
Qwen3.5-flash & \textbf{59.93} & 39.21 & 38.17 \\
\rowcolor{TableStripe}GPT-5.6-Luna & \textbf{79.97} & 65.46 & 68.05 \\
MiniMax-M2.5 & \textbf{60.10} & 42.49 & 38.51 \\
\rowcolor{TableStripe}Kimi-K2.5 & \textbf{64.08} & 53.20 & 48.88 \\
\bottomrule
\end{tabularx}

  \end{minipage}
\end{table*}

\section{Results and Analysis}
\subsection{Answer quality across models}
\RD\ has the highest accuracy in all twelve model and benchmark combinations in Table~\ref{tab:main}. It leads on AMB-Text and LME-KU with all four models. Relative to the better of ACE and ReasoningBank in each setting, the gains range from 1.16 to 19.77 percentage points on AMB-Text and from 6.41 to 25.64 points on LME-KU.

On fact consolidation, \RD\ exceeds the better-performing baseline by 17.00 points with Qwen and 26.00 points with Luna. MiniMax scores 57.00\% with \RD, compared with 42.00\% for ACE and 46.00\% for RB, a gain of 11.00 points over the better-performing baseline. Kimi scores 59.00\%, compared with 41.00\% and 45.00\%, a gain of 14.00 points. The gains are positive across all four backbones, with the largest margin for Luna.

\RD\ has the highest three-suite mean with all four models: 66.59\% for Qwen, 86.71\% for Luna, 77.13\% for MiniMax, and 72.83\% for Kimi. These means weight the three benchmark scores equally; per-benchmark gains show how the size of the advantage varies across tasks.

Fact-consolidation questions require selecting current revisions along a relation path. Curation must extract that path and resolve each relation before packing the context. \RD\ improves fact-consolidation accuracy with all four models.

\subsection{Forgetting and query-conditioned curation}

The $-$FORGET configuration scores 68.60\% versus matched Full's 91.86\% on AMB-Text, 60.26\% versus 93.59\% on LME-KU, and 51.00\% versus 74.00\% on MAB-FC (Table~\ref{tab:ablations}). Relative to matched Full, the deficits are 23.26, 33.33, and 23.00 percentage points. Within the matched Luna configuration, $-$FORGET has the largest score deficit on all three suites, with the largest difference on LME-KU.

\begin{figure*}[t]
  \centering
  \includegraphics[width=\textwidth]{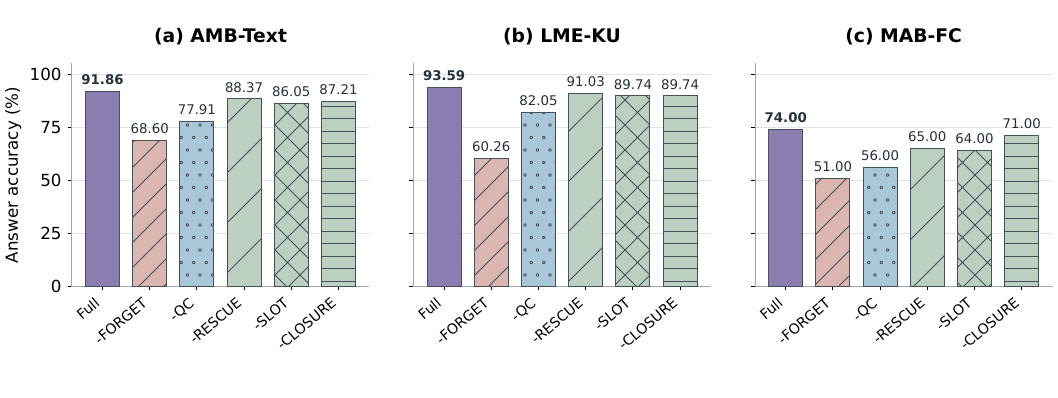}
  \caption{Matched Luna component comparison on (a) AMB-Text, (b) LME-KU, and (c) MAB-FC. Bars show the aggregate answer accuracies from Table~\ref{tab:ablations}; all panels use the same zero-based percentage scale. Purple denotes matched Full. $-$FORGET has the largest deficit relative to matched Full on every benchmark, followed by $-$QC. Variants use the end-to-end protocol in Section~\ref{sec:ablations}.}
  \label{fig:component-comparison}
  \Description{Three panels show the matched Full and five ablation scores. Full scores 91.86, 93.59, and 74.00 percent; minus FORGET scores 68.60, 60.26, and 51.00; minus QC scores 77.91, 82.05, and 56.00. Minus RESCUE, SLOT, and CLOSURE also score below Full on every benchmark.}
\end{figure*}

$-$QC scores 77.91\%, 82.05\%, and 56.00\% on the three suites, respectively. The corresponding deficits relative to matched Full are 13.95, 11.54, and 18.00 points. These are the second-largest deficits among the component ablations on each benchmark. Figure~\ref{fig:component-comparison} contrasts the matched Full and all five variants on a common accuracy scale. A memory formed before seeing the question must anticipate which relations a later answer will require. The $-$QC result occurs despite its larger curation output allowance and query-aware downstream selector.

Across the matched Luna ablations, the score deficit of $-$FORGET relative to matched Full exceeds that of $-$QC by 9.31, 21.79, and 5.00 points on AMB-Text, LME-KU, and MAB-FC, respectively.

\subsection{Evidence chains, slots, and historical eligibility}
$-$CLOSURE scores below matched Full by 4.65 points on AMB-Text, 3.85 points on LME-KU, and 3.00 points on MAB-FC (Table~\ref{tab:ablations}). The 3.00--4.65-point differences are consistent with direct question matches omitting complementary relations needed for some answers.

$-$RESCUE scores below matched Full by 3.49 points on AMB-Text, 2.56 points on LME-KU, and 9.00 points on MAB-FC. The corresponding deficits for $-$SLOT are 5.81, 3.85, and 10.00 points. Both variants score lower on every suite, with the largest differences on MAB-FC. Rescue changes retrieval eligibility for non-active entries; the curator can still recover historical facts directly from the source archive.

\section{Stored History across Query Intents}
\label{sec:extended}
We evaluate the intent-aware extension on BEAM and PersonaMem-v1 with the same four backbones. BEAM uses the \texttt{100K} split, corresponding to the experiment's 128K tier. Its questions cover knowledge updates, contradiction resolution, temporal reasoning, and preference following~\citep{tavakoli2025beam}. PersonaMem-v1 uses the 32K configuration and truncates context at the index specified by each question~\citep{jiang2025personamem}.

The evaluation excludes the first ten development questions from each benchmark and covers 150 BEAM questions and 579 PersonaMem questions per method. BEAM's loader interleaves its four categories within conversation order. Each rubric criterion is graded 0, 0.5, or 1; criterion scores are averaged within a question and question scores are then averaged across the suite. PersonaMem uses the official single-option parsing score over four answer options. Its parser checks the tagged final answer and the full response, preferring parenthesized options. An answer is correct when either extracted option-letter set equals the gold singleton.

\RD\ leads all eight model and benchmark combinations in Table~\ref{tab:extended}. On BEAM, Qwen scores 50.21\%, compared with 32.84\% for ACE and 34.75\% for RB, a gain of 15.46 points over the better-performing baseline. Luna scores 70.58\%, compared with 56.97\% and 53.47\%, a gain of 13.61 points. Kimi scores 57.83\%, exceeding the better-performing baseline by 15.67 points, while MiniMax scores 42.39\%, a gain of 8.47 points over RB. The observed margins span 8.47--15.67 points across the four backbones.

On PersonaMem, Qwen scores 59.93\%, compared with 39.21\% for ACE and 38.17\% for RB, a gain of 20.72 points over the better-performing baseline. Luna scores 79.97\%, compared with 65.46\% and 68.05\%, a gain of 11.92 points. Kimi reaches 64.08\%, ahead of ACE by 10.88 points and RB by 15.20 points. MiniMax reaches 60.10\%, exceeding ACE by 17.61 points and RB by 21.59 points. Thus the advantage over the better-performing baseline ranges from 10.88 to 20.72 points.

The query-intent results extend the comparison to long histories and evolving preferences.

\section{Discussion}
Separating storage from use makes forgetting a decision about which retained observations influence an answer. A superseded fact can leave a current-state context while remaining available for a historical query. Complementary relations can stay useful after another fact is replaced.

The matched Luna ablations have the same ordering on all three suites. Relative to matched Full, $-$FORGET has the largest deficit and $-$QC the next largest. The differences for $-$CLOSURE, $-$RESCUE, and $-$SLOT are smaller.

Fact consolidation requires the evidence view to preserve a usable relation path while resolving revisions within each hop. The retained archive provides the observations; curation determines which relations and versions enter the answer context.

Slot granularity governs this separation in practice. The curator must distinguish complementary facts while grouping competing values of one relation; the materializer checks the same-slot relationship.

\section{Conclusion}
What an agent should forget depends on which stored facts are appropriate to use for the current question. \RD\ implements this separation with a retained source archive and a query-conditioned memory view. The experimental comparisons connect answer quality to query-relevant evidence, revision control, and preservation of complementary relations. The archive preserves availability, while the memory view governs the evidence used for an answer. This separation accommodates both current-state evidence and historical observations whose usefulness changes with the question.

\clearpage
\bibliographystyle{ACM-Reference-Format}
\bibliography{references}

\end{document}